\documentclass[review]{elsarticle}

\usepackage{graphicx}

\usepackage[numbers]{natbib}

\usepackage{tabularx}

\usepackage{float} 

\usepackage{makecell}
\usepackage{multirow}
\usepackage{adjustbox} 
\usepackage{rotating}

\usepackage{subfigure}
\usepackage{color}
\usepackage{lineno,hyperref}

\journal{}

\begin{document}

\begin{frontmatter}

\title{Automatic cephalometric landmarks detection on frontal faces: an approach based on supervised learning techniques}

\author{Lucas Faria Porto}
\address{University of Brasilia - Brazil}
\ead{lucasfporto@gmail.com}
\author{Laise Nascimento Correia Lima}
\address{University of Campinas - Brazil}
\ead{laiselima@gmail.com}
\author{Marta Flores}
\address{University of São Paulo - Brazil}
\ead{mrpflores@gmail.com}
\author{Andrea Valsecchi}
\address{University of Granada - Spain}
\ead{valsecchi.andrea@gmail.com}
\author{Oscar Ibanez}
\address{University of Granada - Spain}
\ead{oscar.ibanez@decsai.ugr.es}
\author{Carlos Eduardo Machado Palhares}
\address{Department of Federal Police - Brazil}
\ead{palhares.cepm@dpf.gov.br}
\author{Flavio de Barros Vidal}
\address{University of Brasilia - Brazil}
\ead{fbvidal@unb.br}




\begin{abstract}
Facial landmarks are employed in many research areas such as facial recognition, craniofacial identification, age and sex estimation among the most important. In the forensic field, the focus is on the analysis of a particular set of facial landmarks, defined as cephalometric landmarks. Previous works demonstrated that the descriptive adequacy of these anatomical references for an indirect application (photo-anthropometric description) increased the marking precision of these points, contributing to a greater reliability of these analyzes. However, most of them are performed manually and all of them are subjectivity inherent to the expert examiners. In this sense, the purpose of this work is the development and validation of automatic techniques to detect cephalometric landmarks from digital images of frontal faces in forensic field. The presented approach uses a combination of computer vision and image processing techniques within a supervised learning procedures. The proposed methodology obtains similar precision to a group of human manual cephalometric reference markers and result to be more accurate against others state-of-the-art facial landmark detection frameworks. It achieves a normalized mean distance (in pixel) error of $0.014$, similar to the mean inter-expert dispersion ($0.009$) and clearly better than other automatic approaches also analyzed along of this work ($0.026$ and $0.101$).
\end{abstract}


\end{frontmatter}


\section{Introduction}
\label{intro}


The use of facial landmarks have been considered in diverse forensic areas such as identification of living individuals~\cite{7025147,lucas2016metric}, age and sex estimation~\cite{palhares2017,BORGES20189,cattaneo2009difficult} or craniofacial identification~\cite{aulsebrook1995superimposition} for the dead. All these areas, landmarks are mainly employed to measure distances, proportions~\cite{Martos18} and to guide image superimposition processes~\cite{Campomanesb2014}. Thus, the performance/reliability of these techniques are strongly influenced by the location of landmarks and they demand precision and repeatability. 

However, in the majority of these scenarios, the landmarks are located following a completely manual approach which makes the whole process subjective and strongly dependent on the combination of three expert's skills: ability, knowledge and experience, as described in~\cite{BORGES20189}. On facial images, the following issues an effect on the accuracy of the landmarks locating: landmark type, facial pose, image resolution, photograph illumination and focus, as described by~\cite{moreton2011investigation}. In fact, as demonstrated in~\cite{campomanes2015dispersion,cummaudo2013pitfalls}, there is a significant landmark location dispersion among experts and also within different landmark location processes from the same expert. In an intermediate situation, the forensic experts from many police authorities daily uses a computational forensic tool. Many of these software\footnote{Following the journal requisites, all information related about (or allow identify) to the authors are hidden for the double-blind reviewing process. In this case, the computational developed tool will be referenced by "software" along the manuscript. Thus, the computational developed tool will be referenced only by "software" along the manuscript.} provide features which assist (or guide) the expert in the identification of most of the cephalometric landmarks. Together with a detailed guide, it demonstrated that is possible to reduce the location dispersion, as presented in~\cite{flores2017manual,flores2014master}. In any case, the repetitive nature of the process, errors related to analyst’s fatigue must be also considered. Additionally, even trained experts need an important amount of time to locate all the required landmarks with high accuracy. 

All the previous issues justify the need for the automation of the landmark location process. The research on automatically identifying and extracting facial features from images, is not a recent scientific field and we easily can found studies from the 80's. However, automatic landmark location in particular have been increasingly attracting the interest of researchers due to its multiple application in fields such as: identification and facial recognition~\cite{wiskott1997face,shi2006effective,8401553}, facial modeling using 3D images~\cite{vezzetti20143d,icinco18Vidal}, tracking~\cite{cech20143d}, cephalometric points identification~\cite{le2014approach}, sex and age estimation~\cite{dibeklioglu2015combining,patil2005determination} and so on. Several studies have been proposed for automatic identification of facial points. Among the various imaging techniques, pattern recognition/computer vision are the most popular~\cite{6751298}. However, all the existing approaches present an important limitation in order to apply them to the forensics field, the set of landmarks recognized by these algorithms are not the cephalometric ones.

In this contribution, we aim to fill in the gap in automatic landmark location by a novel adapting of existing approaches to the particular problem of automatic detection of cephalometric landmarks in frontal face images. This specific objective has been addressed through the following three tasks: i) Analysis of the available automatic algorithms for landmark location; ii) Propose a new methodology especially designed for cephalometric landmarks identification. iii) Performance study of the existing approaches adapted following the proposed methodology and comparison among them and the dispersion of human experts. 

The paper is organized as follows: Section~\ref{sec:material} shows the proposed method and all processing details to identify and extract cephalometric landmarks in the facial image. It also includes a short description of other automatic approaches we have adapted to tackle landmark location. Section~\ref{sec:resultados} is devoted to present the experimental results and comparisons among automatic approaches and experts' manual dispersion. Finally, Section~\ref{sec:conclusao} presents the discussion and main conclusions.


\section{Materials and methods}\label{sec:material}

\begin{figure}[!htpb]
\centering
\includegraphics[width=1\textwidth]{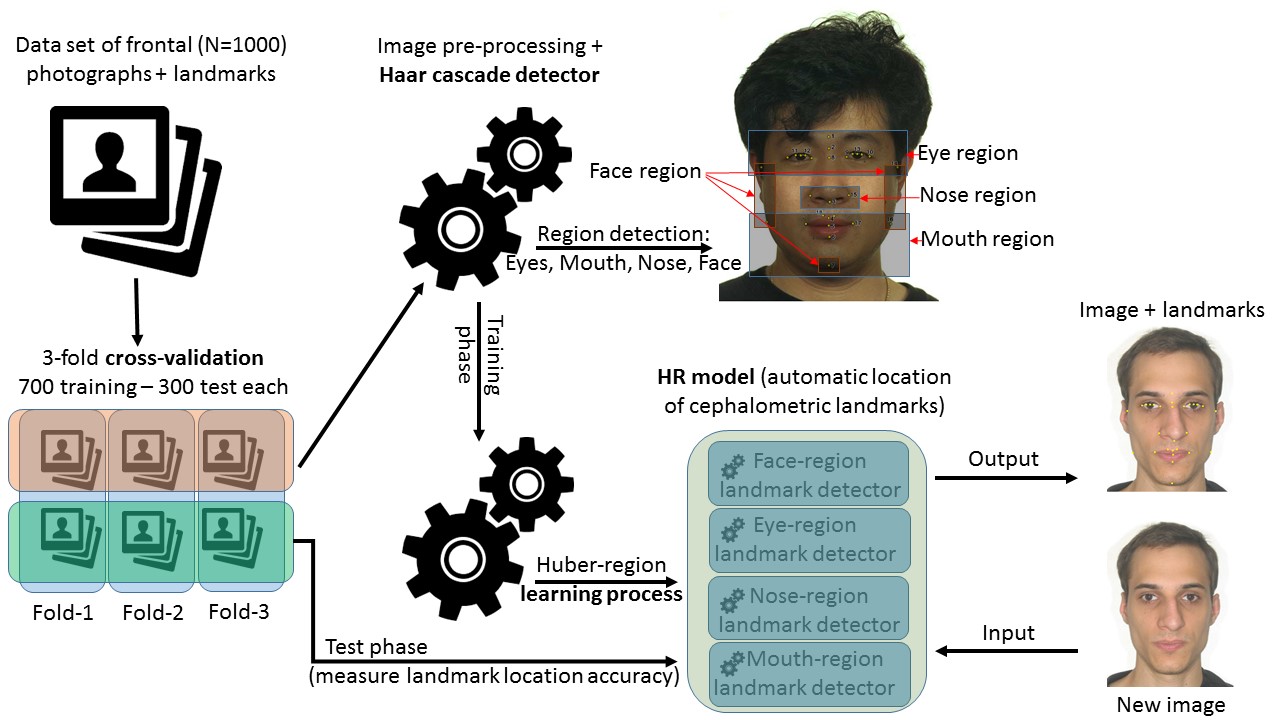}
\caption{Graphical description of the proposed automatic approach and algorithms. It also includes the methodological approach followed for training the learning model and validate it through 3-fold cross validation}
\label{fig:work-flow}
\end{figure}

The algorithm proposed to automate cephalometric landmark location has three main components: image processing, face region detection and training process, as described in Figure \ref{fig:work-flow}. For each face image input a pre-processing step is applied to boost facial features. From this features, the Haar-Cascade~\cite{viola2004robust} face detection technique is able to identify specific bounded regions of interest on the image, first the whole face, and from it, the detected face is divided in the following different regions: eyes, mouth and nose. Finally, a machine learning process is applied for each of the four individual region detected in the previous stage. 

The learning process considers a large set of labeled images (see Section \ref{subsec:database} for a detailed description below), i.e., a data set of facial images together with the location ($x$ and $y$ horizontal and vertical coordinates respectively) of cephalometric landmarks within the image. This learning process is based on the algorithm presented in~\cite{huber2015fitting}. The process followed by the latter consist on identifying a face-bounded region in each photo and all landmark points to create a new training set file. In our proposed methodology, specialized training set files per region of interest are generated using the face area identified in the face region detection stage \footnote{The proposed implementation and the source code is available on: https://github.com/lucasfporto/ACLDTool, all code was developed using C++ language and OpenCV framework~\cite{itseez2015opencv}}. We have called it Huber-Region (HR from now on). In Figure \ref{fig:work-flow} the HR is presented as an input in the learning process methodology proposed in this work.

In the remaining of this Section we present and describe other approaches for automatic landmark location (Subsection~\ref{subsec:autoapro}), the set of images and landmarks used in this work (in Subsection~\ref{subsec:database}) and finally the experimental set-up and metrics employed to validate our approach and to study its reliability in comparison with other automatic proposals and the performance of human experts (Subsection~\ref{subsec:accuracy}).

\subsection{Others automatic facial landmark detection approaches}\label{subsec:autoapro}

As explained previously, automatic landmark location has been addressed in the Computer Vision community for more than twenty years. However, the common handicap all these proposals share is the set of landmarks recognized by these algorithms. They are not cephalometric ones because the recognized landmarks has not been developed for photo-anthropometry analysis, but for face tracking, pose estimation, emotion recognition, face registration and recognition, as described in~\cite{Çeliktutan2013}. 

Among state-of-the art methods two of the most popular approaches are DLIB~\cite{dlib09} and Supervised Descent (SD)~\cite{huber2015fitting,xiong2013supervised}. Both use Histogram of Oriented Gradient (HOG)~\cite{Dalal05} technique to identify the feature of the landmark on a specific position in the image. The DLIB uses HOG in combination with a linear classifier meanwhile the SD combines HOG with the position on the face of each landmark. During the identification process in SD, the algorithm starts looking for the feature on a specific region in the image, making the process fast and avoiding false positive match in another region. 

We have made use of these two algorithms, but in this case, we have trained them with the same image and landmark sets (see Section \ref{subsec:database}) used in our approach, so we can measure and compare the performance of the three automatic approaches in the same conditions.

\subsection{Image and landmark sets}\label{subsec:database}

Our main goal is to automatically identify cephalometric landmarks in frontal face images to be processed for photo-anthropometry analysis. Thus, the set of images employed for training and test the proposed automatic methods is composed of frontal face images where 28 cephalometric landmarks have been manually located by an forensic expert. In particular: the frontal face image of 1,000 individuals was acquired fulfilling the ICAO 9303 normative~\cite{ISO19794-5} for Machine Readable Travel Documents, i.e., travel passport. A camera setup with a 35mm of focal length and the captured subject positioned at 1.5m distance from the camera was used. The images were stored at a spatial resolution of $480 \times 640$ pixels and 24bits of color depth. All captured images were taken over a white and uniform background, neutral facial expression and no glasses. The 1.000 captured samples was equally distributed among male (500) and female (500) for sex class and five different age groups class (6, 10, 14, 18, and 22 years old), each age group class with 200 individuals. The inclusion criteria consisted of Brazilians citizens within the age range of 6-22 years, with neutral facial expression, closed lips and face positioned directly towards the camera. The exclusion criteria consisted of individuals with evident head rotation in the sagittal, axial or coronal planes; as well individuals with evident facial deformities, asymmetries, facial hair, jewelry, make up or those with incomplete photographic registration of the face.

\begin{figure}[!htpb]
\centering
\includegraphics[width=0.7\textwidth]{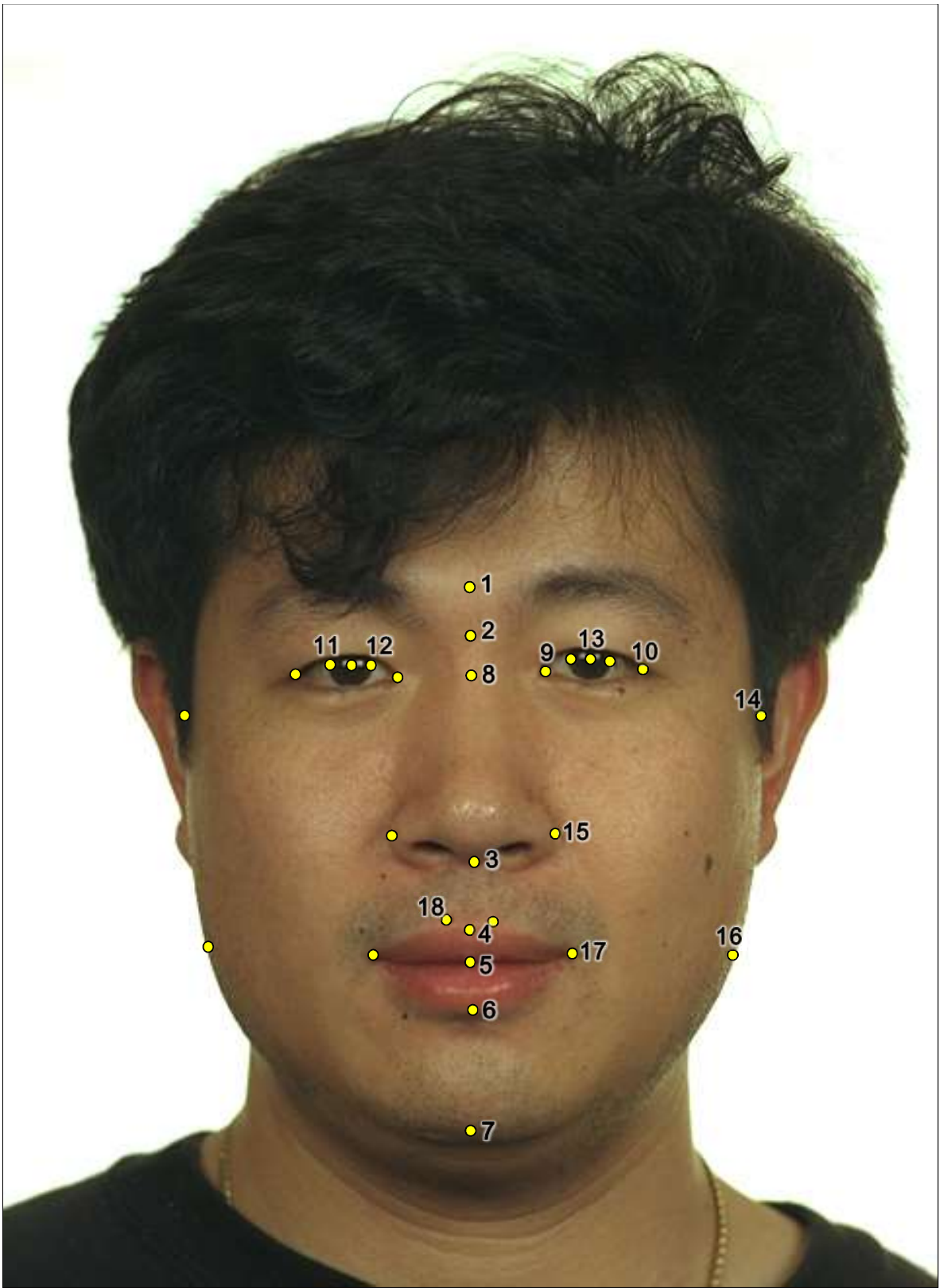}
\caption{All 28 cephalometric landmarks adopted in this work: 1. Glabella (g'-PT); 2. Nasion (n'-PT); 3. Subnasale (sn'); 4. Labiale superius (ls'); 5. Stomion (sto'); 6. Labiale inferius (li'); 7. Gnathion (gn'); 8. Midnasal (m'); 9. Endocanthion (en'); 10. Exocanthion (ex'); 11. Iridion laterale (il); 12. Iridion mediale (im); 13. Pupil (pu'); 14. Zygion (zy'-PT); 15. Alare (al'); 16. Gonion (go'-PT); 17. Cheilion (ch'); 18. Crista philtri (cph'), image adapted from~\protect\cite{phillips2000feret}.}
\label{fig:all_points}
\end{figure}

As described in~\cite{palhares2017} and~\cite{BORGES20189}, one expert manually located 28 cephalometric landmarks in all the 1000 images following the instructions steps described in~\cite{flores2017manual}. The whole list of the selected landmarks, according to Caple and Stephan standard nomenclature~\cite{stephanStandards16}, is the following: \textit{Endocanthion} (en'), \textit{Exocanthion} (ex'), \textit{Iridion laterale} (il), \textit{Iridion mediale} (im), \textit{Pupil} (pu'), \textit{Zygion PT} (zy'-PT), \textit{Alare} (al'), \textit{Gonion PT} (go'-PT) and \textit{Cheilion} (ch'), \textit{Crista philtri} (cph') bilateral landmarks. The remaining can be found on face's midline as: \textit{Glabella PT} (g'-PT), \textit{Nasion} (n'-PT), \textit{Subnasale} (sn'), \textit{Labiale superius} (ls'), \textit{Stomion} (sto'), \textit{Labiale inferius} (li'), \textit{Gnathion} (gn'), \textit{Midnasal} (m')~\cite{brown2004survey}. Figure~\ref{fig:all_points} shows all selected landmarks location on the face, meanwhile the Table~\ref{tab:landmarksCraPto} presents the craniometric and photo-anthropometric landmarks groups, reiterating those landmarks are collected using frontal facial photos resulting in different localization \textit{in vivo}.

\begin{table}[!htpb]
    \centering
    \caption{Craniometric and Photo-anthropometric landmarks}
    \begin{tabular}{ll}
        \hline 
        Craniometric landmarks & Photo-anthropometric landmarks \\ 
        \hline \hline 
        Alare (al')             & Glabella (g'-PT) \\
        Cheilion (ch')          & Gonion (go'-PT) \\
        Crista philtri (cph')   & Nasion (n'-PT) \\
        Endocanthion (en')      & Zygion (zy'-PT) \\
        Exocanthion (ex')       & \\
        Gnathion (gn')          & \\
        Iridion laterale (il)  & \\
        Iridion mediale (im)    & \\
        Labiale inferius (li')  & \\
        Labiale superius (ls')  & \\
        Midnasal (m')           & \\
        Pupil (pu)              & \\
        Stomion (sto')          & \\
        Subnasale (sn')         & \\
        \hline 
    \end{tabular}
    \label{tab:landmarksCraPto}
\end{table}

It is important to notice that four of the previous landmarks were endowed with PT (photo-anthropometric) symbol due to the fact that they had an alternative definition and the used software has specific tools to facilitate its location on frontal photographs (See in second column in Table \ref{tab:landmarksCraPto}). These landmarks and their definition are described in Table \ref{tab:landmarksCraPto}, as follows:

\begin{itemize}
    \item Nasion PT: Intersection of orbital midline with the horizontal line that pass through the middle height of the upper palpebral grooves.
    \item Glabella PT: Intersection between the orbital midline and the horizontal line that intersects the upper edge of the orbital circumferences (automated).
    \item Gonion PT: Most lateral point where the horizontal line of reference passes through Stomion point and crosses the contour line of the face.
    \item Zygion PT: Most lateral point of the face (greater width) of the zygomatic bone.
\end{itemize}

Additionally, two of the previous landmarks (Nasion and Glabella) and two more (Midnasale and Pupila) are automatically located by used software considering the location of other landmarks manually located.



\begin{figure}[!htpb]
\centering
\includegraphics[width=1\textwidth]{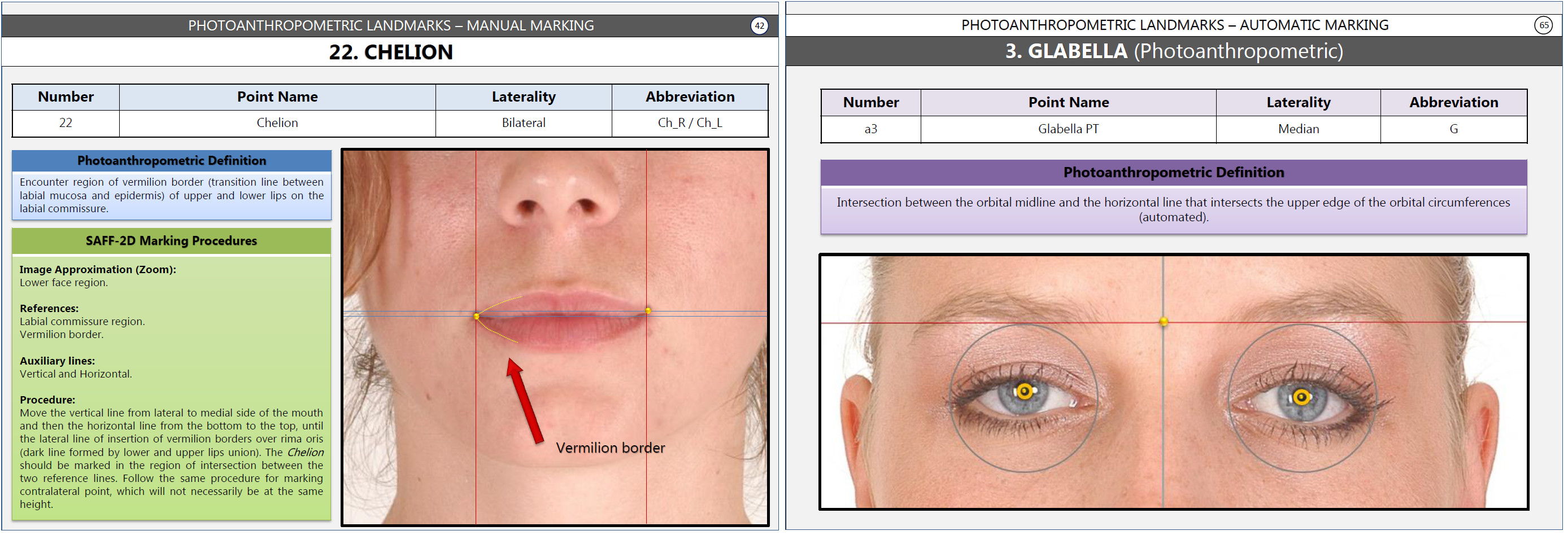}
\caption{Two entries of the software guide of use and best practices. On the left, procedure to locate Cheilion. On the right, automatic procedure followed to locate Glabella PT}
\label{fig:SAFF-2D-guide}
\end{figure}

Together with specialized software, a best practice and an use guide was developed in order to standardize, as much as possible, the landmarks location with the final goal to reduce the inter and intra-expert landmark location dispersion (Figure~\ref{fig:SAFF-2D-guide} shows two entries of the used software guide of use)

Finally, we also aim to measure the accuracy i pixel of our automatic approach in comparison with expert manual location dispersion. For that purpose we have used a different data set composed by 20 images where 12 forensic experts manually located (inter-expert dispersion study) the same set of cephalometric landmarks using the software.  


\subsection{Experimental set-up and evaluation metrics}\label{subsec:accuracy}

As described in Section \ref{sec:material}, we have developed the training process in four specifics regions: face, eyes, mouth and nose. Our complete photo data-set has 1000 images and 1000 ground truth files, each of them containing $x$ and $y$ coordinates of each of the 28 landmarks\footnote{Informed consent was obtained from all individual participants included in the study and all procedures was registered under Brazilian Ethical Committee rules - CAAE 17017213.0.0000.5440.}. 

In order to evaluate the expected performance of a
regression or classifier model over a data set, $k$-fold cross-validation schemes
are commonly used in the machine learning literature~\cite{hastie_09,kohavi1995study}. In our case, a three-fold cross-validation procedure was adopted, randomly separating three times 700 photos for training the detectors and 300 photos for testing. Previous Figure~\ref{fig:work-flow} represents also validation methodology through cross-validation.

As a result of the training process, we get four specialized landmark detectors: Eye landmark detector (it includes Glabella, Nasion, Midnasal, Endocanthion, Exocanthion, Iridion laterale, Iridion mediale, Zygion and Pupil landmark points), Mouth landmark detector (includes Labiale superius, Stomion, Labiale inferius, Cheilion, Gonion, Gnathion and Crista philtri landmark points), Face landmark detector (inclues Zygion, Gonion and Gnathion landmark points) and Nose landmark detector (includes Subnasale and Alare landmark points). 

In order to quantify the algorithm's accuracy, we measure the Euclidean distance in pixels between the landmark location provided by the automatic algorithms and a reference (or ground truth) landmark location, i.e. the $(x,y)$ coordinates pairs of each landmark/photograph according to the manual procedure followed by an expert. In order to allow error comparison (and aggregation) along all the cases (and against future works in the field), we have adopted a normalization process of the Euclidean distances. A mechanism as this is required when comparing distances among images of different spatial resolution and acquisition camera parameters (focal and subject to camera distance). This way, the error metric for a particular pair of landmarks (ground truth and tested one) is the Euclidean distance between them divided by the Euclidean distance between both Exocanthion landmarks.

\section{Results}
\label{sec:resultados}

Firstly, we performed a comparative analysis between the proposed algorithm and the two automatic framework counterparts, DLIB~\cite{dlib09} and SD~\cite{huber2015fitting,xiong2013supervised}. According to Subsection \ref{subsec:accuracy} each algorithm was tested on 900 images (three test sets composed of 300 images each) with 28 landmarks each. For each individual landmark/image, 25.200 landmarks was used in total, we calculate the Euclidean Distance (in pixels) between the location provided by each automatic algorithm and the location provided by the expert (ground truth). Using the resulting value (distance error) we ranked each algorithm from best performing to worse and finally calculated the Average ranking, presented in Table~\ref{tab:900_mean_ranking}. According to the mean ranking, the proposed algorithm (HR) clearly outperforms the two other automatic approaches: ranking 1 against 2.1 (SD) and 2.9 (DLIB). 

\begin{table}[!htpb]
\caption{Average ranking by algorithm.}
\centering
\begin{tabular}{rlr}
  \hline
    & Algorithm & Average Ranking \\ 
  \hline
  1 & HR & 1.0 \\ 
  2 & SD & 2.1 \\ 
  3 & DLIB & 2.9 \\ 
   \hline
\end{tabular}
\label{tab:900_mean_ranking}
\end{table}

To evaluate the statistical significance test, of the previous ranked algorithms described in Table \ref{tab:900_mean_ranking}, we have performed a Non-parametric Wilcoxon Signed-rank test~\cite{WilcoxonTestRef} using all 25.200 ranking values for each algorithm. We have obtained the same $p$-value of~$7.6\times10^{-6}$ for the HR vs. SD and HR vs. DLIB, what certifies statistically the significant superior performance of HR.

Looking closer to the algorithm's performance, the Table~\ref{tab:900_detailed_algo} shows the average distance of each automatic approach for each cephalometric landmark. As expected, HR performs always better than SD and DLIB, with large differences in the majority of the cases. HR mean distance error lies between $0.010$ and $0.019$ for all landmarks a part from both Zygion, were the error increases to $0.031$.

\begin{table}[!htpb]
\centering
\caption{Mean distance error by automatic approach and landmark.}
\begin{tabular}{ccc|ccc}
  \hline 
  Landmark & Algorithm & Mean Distance & Landmark & Algorithm & Mean Distance \\ 
  \hline 
  \multirow{3}{*}{al'}
   & HR & 0.012 & \multirow{3}{*}{ch'} & HR & 0.014 \\ 
   & SD & 0.027 & & SD & 0.029 \\ 
   & DLIB & 0.177 & & DLIB & 0.024 \\ \hline
  \multirow{3}{*}{cph'} 
   & HR & 0.014 & \multirow{3}{*}{ex'} & HR & 0.013 \\
   & SD & 0.024 & & SD & 0.023 \\ 
   & DLIB & 0.027 & & DLIB & 0.028 \\ \hline 
  \multirow{3}{*}{en'} 
   & HR & 0.019 & \multirow{3}{*}{g'-PT} & HR & 0.012 \\
   & SD & 0.029 & & SD & 0.026 \\   
   & DLIB & 0.139 & & DLIB & 0.168 \\  \hline 
  \multirow{3}{*}{gn'}
   & HR & 0.012 & \multirow{3}{*}{go'-PT} & HR & 0.011 \\
   & SD & 0.020 & & SD & 0.019 \\   
   & DLIB & 0.035 & & DLIB & 0.027 \\ \hline 
  \multirow{3}{*}{il}
   & HR & 0.012 & \multirow{3}{*}{im} & HR & 0.015 \\ 
   & SD & 0.019 & & SD & 0.025 \\   
   & DLIB & 0.052 & & DLIB & 0.034 \\ \hline 
  \multirow{3}{*}{li'} 
   & HR & 0.016 & \multirow{3}{*}{ls'} & HR & 0.012 \\ 
   & SD & 0.025 & & SD & 0.023 \\   
   & DLIB & 0.032 & & DLIB & 0.165 \\ \hline 
  \multirow{3}{*}{m'} 
   & HR & 0.012 & \multirow{3}{*}{n'-PT} & HR & 0.015 \\
   & SD & 0.021 & & SD & 0.026 \\   
   & DLIB & 0.143 & & DLIB & 0.035 \\ \hline
  \multirow{3}{*}{pu'} 
   & HR & 0.010 & \multirow{3}{*}{sn'} & HR & 0.010 \\ 
   & SD & 0.022 & & SD & 0.022 \\  
   & DLIB & 0.163 & & DLIB & 0.154 \\ \hline
  \multirow{3}{*}{sto'} 
   & HR & 0.010 & \multirow{3}{*}{zy'-PT} & HR & 0.031 \\
   & SD & 0.021 & & SD & 0.055 \\   
   & DLIB & 0.158 & & DLIB & 0.161 \\
  \hline 
\end{tabular}
\label{tab:900_detailed_algo}
\end{table}

Secondly, we also aim to measure the accuracy of the automatic approaches with special focus in to HR, our proposal and best performing automatic approach, for comparison with expert's manual location dispersion. In this purposes, we use a different data set composed by $20$ images where $12$ forensic experts manually located by an inter-expert dispersion study, of the same set of cephalometric landmarks using the software. Table~\ref{tab:inter_experts_vs_algo} shows, in the first row, the mean normalized error distance of the 12 forensic experts on the reduced set of 20 images. The following rows refer to the mean normalized error distance of the three automatic approaches. 

\begin{table}[!htpb]
\centering
\caption{General result: Experts~(\textit{inter} observer) vs automatic frameworks. The lower values represent that the cephalometric landmarks are concentrated near each other meaning greater accuracy.}
\begin{tabular}{rlr}
  \hline
 & Algorithm & Mean Distance \\ 
  \hline
  1 & Expert & 0.009 \\ 
  2 & HR & 0.014 \\ 
  3 & SD & 0.026 \\ 
  4 & DLIB & 0.101 \\ 
   \hline
\end{tabular}
\label{tab:inter_experts_vs_algo}
\end{table}

The mean error of the 12 forensic experts is $0.009$, while the mean error of our approach (HR) is 0.014. SD (0.026) and specially DLIB (0.101) performance is far away from the two previous.

\begin{figure}[!htpb]
\centering
\includegraphics[width=\textwidth]{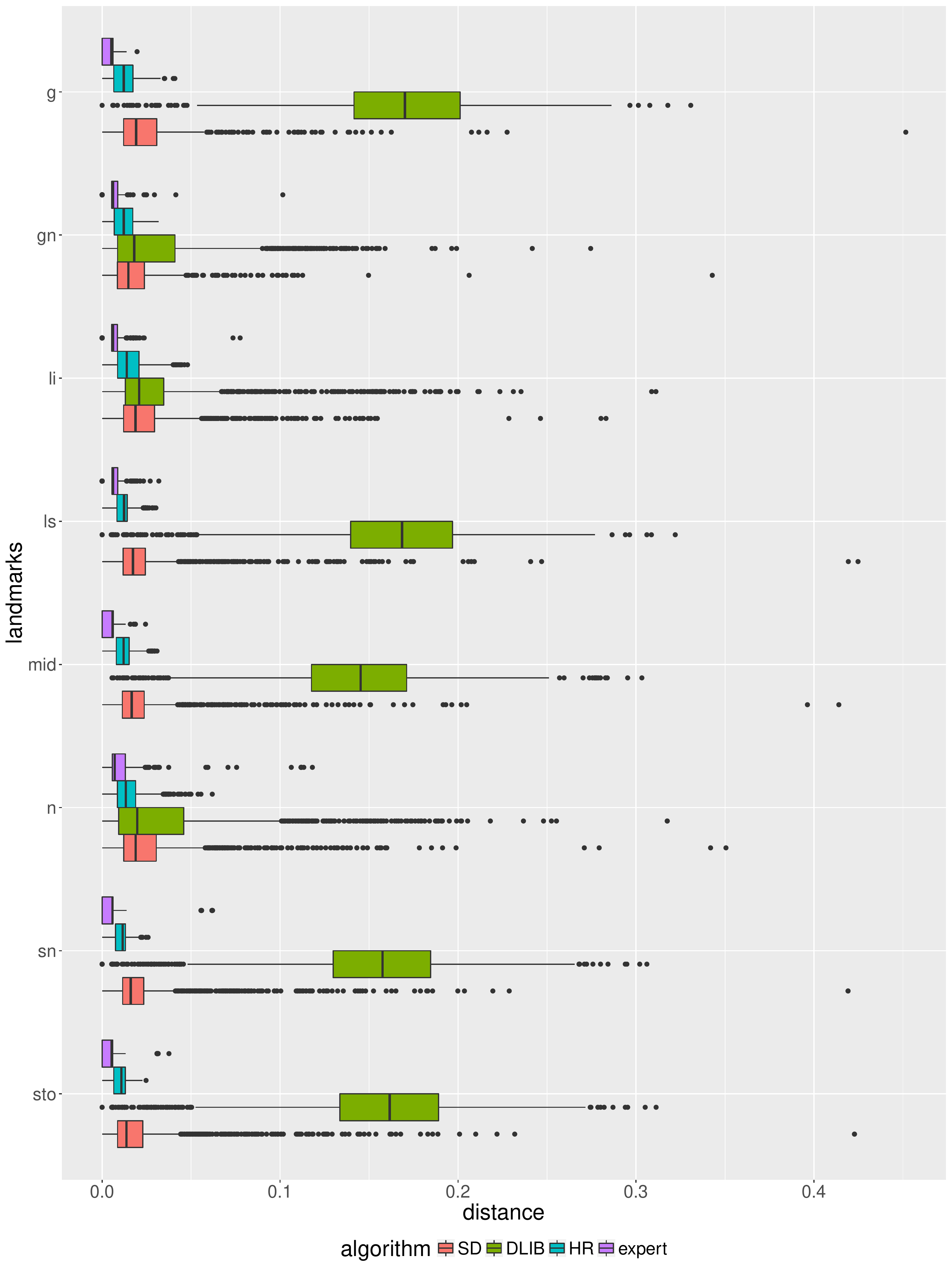}
\caption{Box-plots per each median landmark and location approach (the three automatic methods and the inter-expert dispersion).}
\label{fig:inter_experts_vs_algo_medial}
\end{figure}

\begin{figure}[!htpb]
\centering
\includegraphics[width=\textwidth]{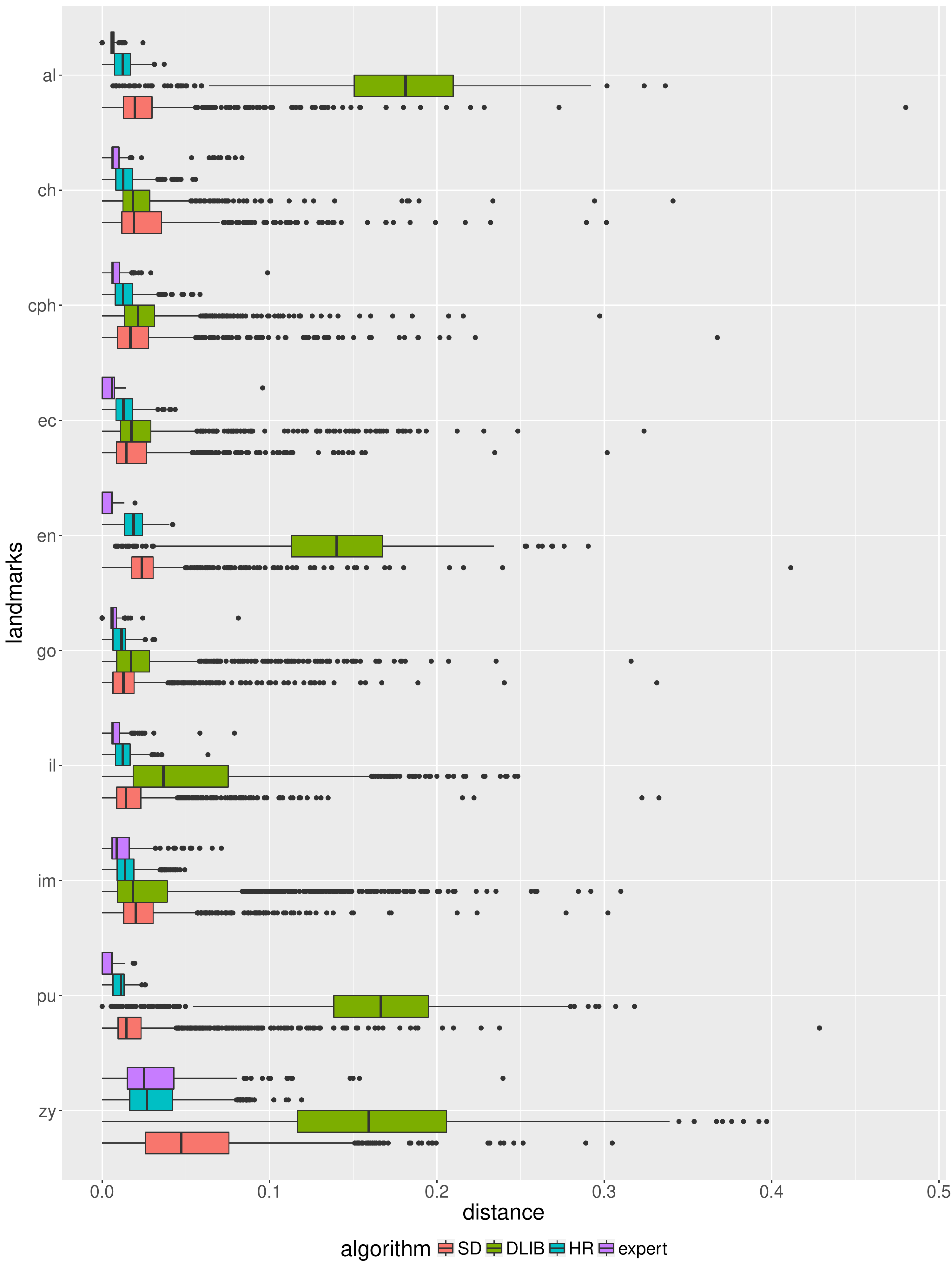}
\caption{Box-plots per each bilateral landmark and location approach (the three automatic methods and the inter-expert dispersion).}
\label{fig:inter_experts_vs_algo_bilat}
\end{figure}

In Figure~\ref{fig:inter_experts_vs_algo_medial}~(median landmarks) and in Figure~\ref{fig:inter_experts_vs_algo_bilat}~(bilateral landmarks), we show box-plots summarizing the performance of the three automatic approaches, over the large data set with three-fold cross validation with 300 images each fold and the inter-expert with 12 forensic experts location dispersion on the reduced data set analyzing 20 images. Although, the comparison between automatic approaches and inter-expert dispersion is based on different data sets (900 vs. 20), it allows us to see how close our automatic approach is from the inter-expert dispersion.  

\begin{figure}[!htpb]
\centering
\includegraphics[width=1\textwidth]{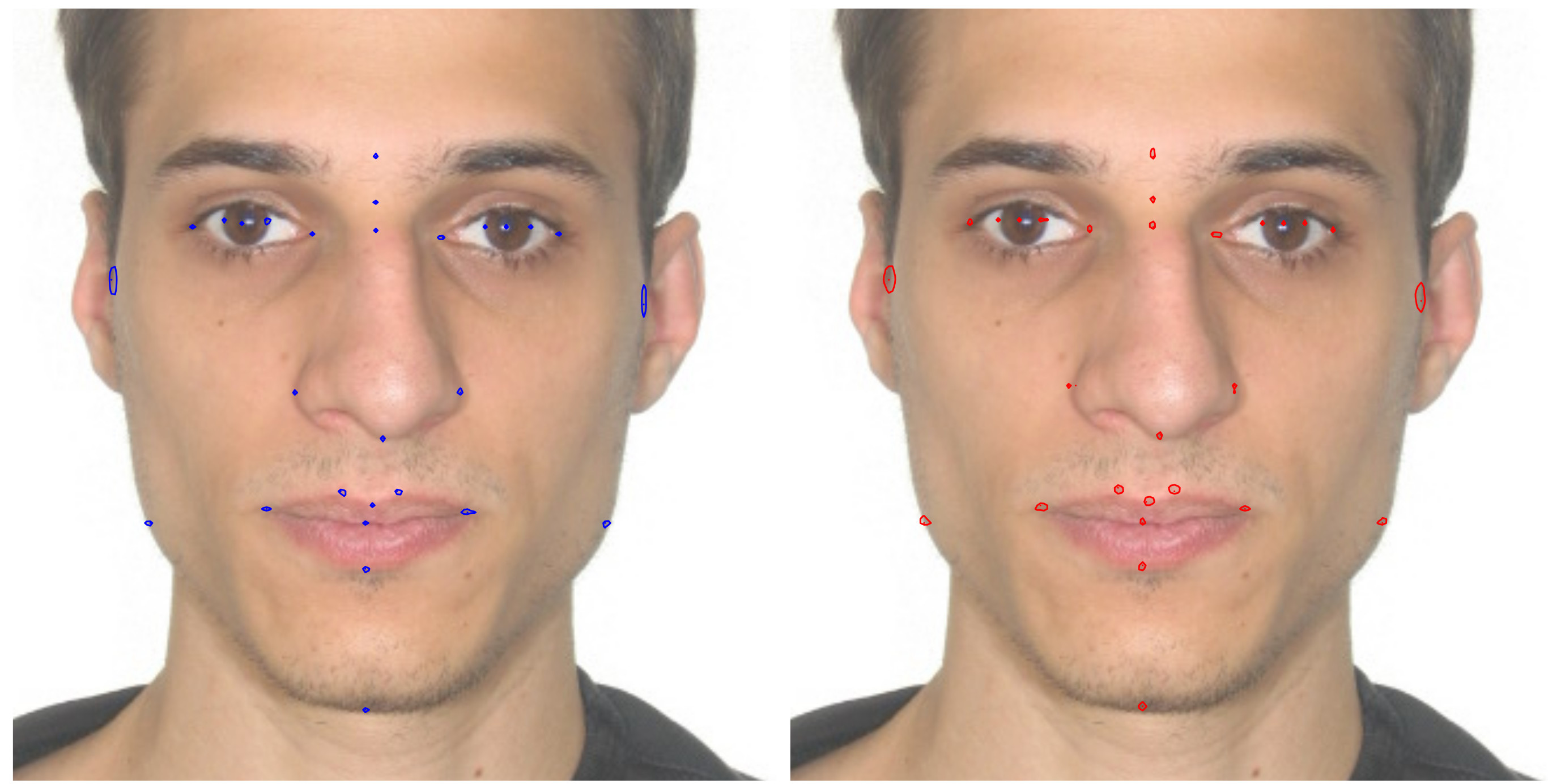}
\caption{The spatial distribution of the located landmark performed by experts (left) and our automatic method (right). For each landmark, we plotted the areas corresponding to the 50\% and 100\% of the distribution, as estimated through kernel density estimation. The 50\% area is so small that is hardly visible.}
\label{fig:face-lm-disper-exp-algo}
\end{figure}

Finally, the Figure~\ref{fig:face-lm-disper-exp-algo} allows a visual comparison of the results of our proposed method against inter-expert variability. On the top of the sample photograph shows the spatial distribution of the landmarks evaluated over all data set photos. The size and shape of the area around each facial landmark show the accuracy of the located landmark performed by our algorithm (right) and the variability among human annotations (left). While inter-expert variability is lower, the automatic landmark location is just slightly less precise and the overall error is very small, if compared to the size of the face.

\section{Discussions}\label{sec:conclusao}

In this manuscript, we have addressed an automatic location of cephalometric  landmarks, a key element for several forensic anthropology tasks including: photo-anthropometry of living useful for forensic identification, sex and age estimation and capable to be adapted to dead people in craniofacial identification (reconstruction and superimposition).  

We have proposed a methodology based on Supervise Learning and one specific algorithm that was trained and tested on a set of 1000 frontal face images, using 28 landmarks were previously locate by one forensic expert. Additionally, we have adapted two popular methods for automatic landmark detection and compare their performance considering also landmark location dispersion by human experts (12 experts on 20 images).

The results show that the proposed methodology clearly overcomes the two counterparts in statistical significant differences and its performance is really close to human location dispersion (0.014 vs 0.009 mean distance error). 

As expected, zygion, a Type 3 landmark, in according to~\cite{Bookstein91}, shows a large dispersion. Both in the locations provided by our HR method and in the case of the inter-expert study. Even a slightly greater variability in the case of human locations (see Figure~\ref{fig:inter_experts_vs_algo_medial} and Figure~\ref{fig:inter_experts_vs_algo_bilat}). Contrary to existing studies on human landmark location dispersion, described in~\cite{campomanes2015dispersion,cummaudo2013pitfalls}, the dispersion location of other Type 3 landmarks was really low, as in the case of gnathion, alare, gonion and glabella. Again, this dispersion was similar in both approaches: automatic and manual (12 experts), what contradicts previous studies where the dispersion in gonion~\cite{campomanes2015dispersion,cummaudo2013pitfalls} and gnathion~\cite{campomanes2015dispersion} was really high, and even in alare with an intermediate dispersion in~\cite{campomanes2015dispersion}. A possible explanation of this phenomena is the fact that the software employed by these authors, facilitates landmark location providing tools such as horizontal and vertical mobile reference lines. On the other hand, the employed gonion and glabella were endowed with PT (photo-anthropometric) symbol due to the fact that they had an alternative definition, and the software has specific tools to facilitate their location on frontal photographs. Thus, the fact of using the software for landmark location seems to involve a lower location dispersion of the landmarks manually located by human experts and even-though, the performance of our automatic method is similar.  

After all, we can conclude that our automatic method is accurate and robust to locate cephalometric landmarks in frontal faces. Currently, in the Brazilian Federal Police Department, the forensic expert's agents use a manual approach (used software in this manuscript) to collect facial cephalometric landmarks on photos. The development of the proposed methodology, will allow them to achieve great improvements in all manual tasks, reducing failures in forensics analysis where human error could compromises the entire forensic analysis. In addition, it will allow them to study the several millions of facial images they have in their data base. 

However, there is an important limitation of the method provided in this work, it has been only trained and tested on frontal facial images fulfilling ICAO 9303 normative~\cite{ISO19794-5}. Future research works should be addressed to tackle and/or test automatic approaches on images showing different poses and acquisition conditions. A different interesting future work would be to analyze if the accuracy/dispersion of the current method is enough for other recent automatic methods in forensic anthropology such as facial examination~\cite{Martos18}, skull-face anatomical/morphological correspondence~\cite{Campomanes18} or skull-face overlay~\cite{Valsecchi18}.

A final future research line, which we would like to address, is the application of the current state-of-the-art machine learning methods, deep learning~\cite{Lecun15}, to the same problem addressed in this article. The deep learning approaches, the Convolutional Neural Networks (CNN) in particular, have revolutionized the field of Computer Vision. Serve as an example DeepFace~\cite{DeepFace14}, a CNN developed by Facebook researchers which achieved human level results in Face Verification task, or HyperFace~\cite{HyperFace} which managed to automatically locate landmarks (not cephalometric ones), the pose of the face and the sex of the individual ``in the wild'' images.


\section*{References}

\bibliography{ref.bib}

\end{document}